\documentclass[preprint,12pt]{article}
\usepackage{graphicx}
\usepackage{algpseudocode}
\usepackage{algorithm}
\usepackage{amsmath}
\usepackage{amssymb}
\usepackage{mathtools}
\usepackage{multirow}
\usepackage{lipsum}
\usepackage{comment}
\usepackage{xcolor}
\usepackage{caption}
\usepackage{subcaption}
\usepackage[affil-it]{authblk}
\usepackage{cite}

\title{Graph-based methods coupled with specific distributional distances for adversarial attack detection}

\author[1,2]{Dwight Nwaigwe}
\affil[1]{Univ. Grenoble Alpes, Inria, CNRS, LJK. Grenoble. France}

\author[1,2]{Lucrezia Carboni}

\author[3]{Martial Mermillod}
\affil[3]{Univ. Grenoble Alpes, Univ. Savoie Mont Blanc, CNRS, LPNC. Grenoble. France}

\author[1]{Sophie Achard}

\author[2]{Michel Dojat}
\affil[2]{Univ. Grenoble Alpes, Inserm, U1216, Grenoble Institut Neurosciences. Grenoble. France}

\begin{document}
\maketitle

\begin{abstract}
Artificial neural networks are prone to being fooled by carefully perturbed inputs which cause an egregious misclassification. These \textit{adversarial} attacks have been the focus of extensive research. Likewise, there has been an abundance of research in ways to detect and defend against them. We introduce a novel approach of detection and interpretation of adversarial attacks from a graph perspective. For an input image, we compute an associated sparse graph using the layer-wise relevance propagation algorithm \cite{bach15}. Specifically, we only keep edges of the neural network with the highest relevance values. Three quantities are then computed from the graph which are then compared against those computed from the training set. The result of the comparison is a classification of  the image as benign or adversarial.  To make the comparison, two classification methods are introduced: 1) an explicit formula based on Wasserstein distance applied to the degree of node and 2) a logistic regression.  Both classification methods produce strong results which lead us to believe that a graph-based interpretation of adversarial attacks is valuable.

\end{abstract}

\resizebox{1.2\textwidth}{!}{ 
{\centering
\includegraphics{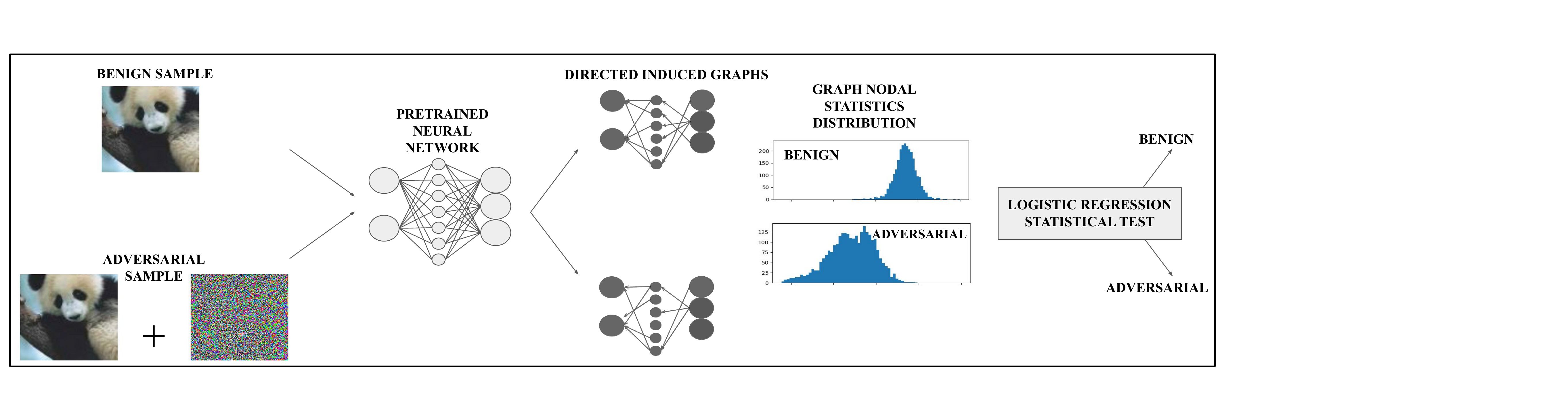}
\par
}
}

\section{Introduction}
\label{introduction}

Artificial neural networks (ANN) are known to be prone to misclassifying carefully perturbed inputs \cite{goodfellow}. These perturbed inputs, called adversarial, have been at the forefront of research in the machine learning community for the past decade. There is a lot of interest in creating new adversarial detection and defense methods, especially as this has consequence for a variety of real-world domains that rely on ANN for classification  \cite{CIRESAN12}, \cite{finlayson19}, \cite{xu19}.

But among the known methods it is apparent that few of them, as diverse as they are, study adversarial attacks from a graph theory perspective. The objective of this paper is the exploration of adversarial attacks using graph-based methods. Indeed, the ANN structure can be described by a graph. In the most basic example, if one considers a standard feedforward ANN, then in a graphical representation, the neurons are associated to vertices/nodes and the weights between them are associated to edges. One may take this representation as inspiration for studying ANN from a graph perspective, although we stress that there is more than one way to obtain a graph from an ANN. 

In \cite{hassabis}, the authors provide a survey of the history of interactions between neuroscience and artificial intelligence and they note how much of the modern success in artificial intelligence can be traced to the understanding of or inspiration by biological systems. There is a line of research in neuroscience that studies the brain using elements of graph theory \cite{bullmore}, and this provides some motivation for the use of graph-theoretic approaches to studying ANN. This can be seen as part of a wider effort in research that seeks to improve artificial intelligence using bio-inspired ideas called \textit{neuromorphic} computing. More examples from neuromorphic computing are \cite{Yang123}, \cite{Yang392}, \cite{Yang930}, \cite{Yang672}. Bio-inspired methods seek to bring the benefits of low-power consumption and more efficient performance or reliability by comparison and inspiration from the human brain. 

In this document, we study the detection of adversarial examples using graphs. Given an input to the neural network, we compute an associated sparse graph. From this graph, we then use a combination of three indices: the most relevant outermost edges, the degree of nodes, and an importance measure. These indices are then used to predict whether the input is adversarial. For this purpose, we test two very different  approaches- a logistic regression and a statistical test based on the Wasserstein distance applied to the degree of nodes.  Lastly, we interpret the relative strength of attacks through our graph-based approach. An advantage of the detection methods introduced is that they include a thresholding step which is non-differentiable, thereby precluding gradient masking \cite{grad-masking} and making it difficult to make adaptive attacks. As part of our studies we also provide benchmarks.

\section{Background and related work}

There have been some efforts in interpreting ANN in graph-theoretic ways. The authors of  \cite{zambra} study the existence and properties of \textit{motifs}, clusters of neurons in ANN which appear often. In \cite{lamalfa}, the authors interpret ANN as a graph and study how MNIST and CIFAR datasets exhibit different distributions under defined quantities (e.g. node input strength, neuron strength). In \cite{corneanu}, a topological study of ANN is made via its \textit{functional graph}, a graph obtained from correlations of neurons. Other work \cite{mocanu},\cite{liu2020},\cite{Naitzat} apply a similar topological view to studying ANN. Despite relating graphs to ANN, none of these articles demonstrate how to use graphs to detect adversarial examples, nor do they provide statistics on detection. An interesting use of graphs occurs in \cite{cheng} where they are used to evaluate the robustness of an ANN, as opposed to adversarial detection. In \cite{xingjun} (``LID"), \cite{kimin}, \cite{harder}, and \cite{feinman}, logistic regression is used to classify an input as benign or adversarial based on certain features, none of which are graph related. The \textit{Local Intrinsic Dimensionality} (LID) of a sample is introduced in \cite{houle} and is motivated by the growth of the volume of a ball as a function of radius and dimension. In \cite{houle}, an analogy is made by replacing volume with a cumulative probability distribution.  The premise of LID is that adversarial examples exhibit significantly greater LID scores than do benign examples.  Given a sample input to a pretrained neural network, the LID method begins by computing the neural network's activations. For each activation, an approximation to its LID score is made, and the result is given to a logistic classifier to determine if the sample input is benign or adversarial. Statistical approaches to adversarial detection can be found in \cite{drenkow} (``RSA") and \cite{roth19}, also neither of which use graph methods. In Random Subspace Analysis (RSA) \cite{drenkow}, projections are used to lower the dimensionality of the space of the pretrained neural network's activations before classifying an input as benign or adversarial. For each layer of the pretrained neural network, a set of approximate projection matrices are created to project the dimension of each layer onto a much smaller space. Then, class protoypes are defined for each class by obtaining the average of each neuron's activations.  For a given input to the neural network that one wishes to classify as benign or adversarial, the following is done: for a given layer and for each projection matrix, compute the distance in projected space between the class prototype and the sample input's activations; record the class that minimizes this difference and add it to a set. After applying this procedure for each layer, one obtains a set of sets where the first set is indexed by layer and the second is indexed by the projection matrix. From this set of sets, a type of voting procedure is done and if the result is less than a threshold, the sample input is labeled adversarial. Our methods extend and complement the previous methods by showing the power of graph theory perspectives applied to studying adversarial attacks, from either a logistic regression or a pure statistics perspective. We also compare our methods with LID and RSA.

\section{Graph generation and quantities of interest}
\label{sec:graph_generation}

To compute the associated directed graph $\mathcal{G}$ for a neural network and input pair, we use layer-wise relevance propagation \cite{bach15}, \cite{bach_chapter}. This algorithm produces a saliency map, but its intermediate steps allows one to assign quantities to neurons which can be interpreted as an indicator of the influence that a neuron has on the output.  We begin by providing a brief overview of layer-wise relevance propagation. Following the notation in \cite{bach_chapter} for the LRP-$\alpha \beta$ rule, signals are propagated from the output layer towards the input layers. For a neuron $k$ in layer $\ell+1$ that is connected to neuron $i$ in layer $\ell$, the propagated signal from $k$ to $i$ is defined to be
\begin{equation}
\label{eq:lrp}
R_{i,k}^{\ell, \ell +1 }= R_{k}^{ \ell +1 } \left( \alpha\frac{a_i \max( w_{ik},0)}{\epsilon+ \sum_h a_h  \max( w_{hk},0)  } - \beta \frac{a_i \min( w_{ik},0)}{\epsilon+ \sum_h a_h \min(w_{hk},0)  } \right)
\end{equation}
where $R_{k}^{ \ell }$ is the relevance of neuron $k$, $a_i$ is the activation of neuron $i$ in layer $\ell$; $w_{hk}$ is the weight between neurons $h,k$; $\epsilon$ is a small parameter; and $\alpha-\beta=1$.  The relevance of a neuron $k$ in layer $l$ is given by 
\begin{equation}
\label{eq:lrp2}
R_{k}^{\ell }= \sum_i R_{k,i}^{\ell, \ell +1 }.
\end{equation}
Equations \eqref{eq:lrp} and \eqref{eq:lrp2} are used together recursively. To start the algorithm, one assigns the relevance of the output neurons of the neural network to be equal to the neural network output. Upon completion of the algorithm, we rank the edge relevance scores $\{R_{i,k}^{\ell, \ell +1 }\}$  in descending order and keep the top 1\%. Our thresholding is inspired by \cite{bullmore}, {\cite{sporns}. In these works, the authors only keep the strongest connections.} The edges that are kept become the edges in our induced graph $\mathcal{G}$. One can compute various quantities from $\mathcal{G}$. One such quantity is given by 

\begin{equation}
\label{eq:modcc}
    I(v_i)= \sum_{j \neq i} \frac{1}{2^{d(v_i,v_j)}}
\end{equation}
where $\{v_i\}_i$ is the set of nodes and $d(v_i,v_j)$ is the distance between vertices $v_i,v_j$. We note that the distance between adjacent nodes is given by \eqref{eq:lrp} and the distance between any pair of nodes is given by the path that corresponds to the shortest sum of distances of adjacent nodes. An intuitive meaning of \eqref{eq:modcc} is that more importance is given to a vertex that has many neighbors and short distances to them. This equation is inspired by closeness centrality \cite{bavelas} which is given by

\begin{equation}
\label{eq:cc}
    C(v_i)=  \frac{1}{\sum_{j \neq i} d(v_i,v_j)}.
\end{equation}
A difference between \eqref{eq:modcc} and \eqref{eq:cc} is that the former is monotone in the cardinality of $\{v_i\}_i$. For bipartite graphs, or ``stacks" of bipartite graphs (one can think of multi-layer perceptrons) a measure of closeness centrality tends to be not useful, hence the motivation for \eqref{eq:modcc}. 

Another quantity of interest is the degree of a vertex,  which is defined to be the difference between out degree and in degree:
\begin{equation}
\label{eq:deg}
    \deg(v) = \deg_{out}(v)-\deg_{in}(v).
\end{equation}
Our last quantity of interest are the values of certain edges of  $\mathcal{G}$. This allows us to incorporate some of $\mathcal{G}$'s topology.  The edges we use are those that correspond to the last two layers of the original neural network. We only use these edges because using all edges would require a data structure of size $O(n_1 n_2, ... n_{l})$, where $n_i$ is the number of nodes in layer $i$. Clearly, this requires an extensive amount of memory when $n_i$ is large.  One can see that in general, when using graph data, it is preferable, at least from a memory standpoint, to use a quantity whose size is  much smaller than $O(n_1 n_2, ... n_{l})$, for instance a dataset whose size is $O(|V|)$, where $V$ is the set of nodes. In fact, our use of degree and node importance as computed for each node meets this constraint.

In \cite{guo},  the neurons just before the softmax layer are studied, which has a similarity with our study of outermost edge relevance. In that article, the authors use the said neurons to compare robustness of non-human primate and human vision with regards to adversarial images. This lends a further (biological) motivation to our use of outermost edge relevance. Since we apply a threshold to the edges of $\mathcal{G}$, there are nodes of $\mathcal{G}$ which are not adjacent to an edge. More generally, the edges among the set $\{ \mathcal{G}_i\}_i$ need not be the same, where $\{ \mathcal{G}_i\}_i$ represents a set of graphs induced from the same architecture. To enforce consistency of representation for the outermost edge relevances we create a weighted adjacency matrix of the same dimension as the adjacency matrix for nodes in the last two layers. The edge relevance values that are above the threshold are recorded as is and those below this percentile are set to 0. The matrix is then flattened into a vector. This flattened vector is our third quantity of interest, and its nonzero components are given by \eqref{eq:lrp}, assuming the component is greater than the threshold.

Lastly, we note that it would be very difficult to create an adaptive attack to counter the methodology proposed here since our detection methods involve graph thresholding, a non-differentiable operation.

\begin{table}[H]
\caption{Summary of relevant graph statistics. } 
\label{table:graph_stat_summary}
\vskip 0.1in
\begin{center}
\begin{small}
\begin{sc}
\resizebox{1\textwidth}{!}{ 
\begin{tabular}{||c c  ||} 
 \hline  \hline
 formula  & name   \\ [0.5ex] 
 \hline
 $R_{i,k}^{\ell, \ell +1 }=  \text{threshold}\left( R_{k}^{ \ell +1 } \left( \alpha\frac{a_i \max( w_{ik},0)}{\epsilon+ \sum_h a_h  \max( w_{hk},0)  } - \beta \frac{a_i \min( w_{ik},0)}{\epsilon+ \sum_h a_h \min(w_{hk},0)  } \right) \right) $& outermost edge relevance  \\ [0.7ex] 
 \hline
 $ I(v_i)= \sum_{j \neq i} \frac{1}{2^{d(v_i,v_j)}}$ & node importance  \\ [0.7ex] 
  \hline
  $\deg = \deg_{out}(v)-\deg_{in}(v)$ & degree  \\ [0.7ex] 
 \hline
\end{tabular}
}
\end{sc}
\end{small}
\end{center}
\vskip 0.1in
\end{table}

\section{A statistical test based on Wasserstein distances}

The Wasserstein-1 distance between two probability distributions $p$ and $q$ defined on a measurable metric space $\mathcal{X}$ is given by

\begin{equation}
\label{eq:wassestein}
\mathcal{W}(p,q) = \min_{\pi(x,y) \in \Pi} \int \left \lVert x-y \right \rVert_1 d\pi(x,y)
\end{equation}
where $\Pi$ is the set of all measures on $(\mathcal{X}, \mathcal{X})$ whose marginal distributions are given by $p$ and $q$. In the case when $p$ consists of one sample $x$ and $q$ consists of discrete samples $(y_i)_{i=1}^N$, then 
\begin{equation}
\label{eq:wassestein:discrete}
\mathcal{W}( \delta_{x}, q)= \frac{1}{N} \sum_i^N \| x-y_i \|_1.
\end{equation}
where $\delta_x$ is the distribution with support at $x$.  Wasserstein distances have been applied to machine learning in several ways. In \cite{cherian}, Wasserstein distances are used to compress data into a small dimensional subspace while maintaining a large distance from adversarial distributions. Other work \cite{wong} uses Wasserstein distances to create adversarial attacks. 

Our goal in using Wasserstein distances is different than those in the examples mentioned. Our goal is to apply Wasserstein differences for benign and adversarial graph statistics in order to classify an input as benign or adversarial. The statistic we are concerned with is degree.

Let $ \hat{\mathcal{B}}_i$ denote the empirical distribution of degree in the case when benign inputs are correctly classified as belonging to class $i$. Similarly, let $\hat{\mathcal{A}}_i$ denote the empirical distribution that corresponds to perturbed inputs which the model incorrectly classifies as belonging to class $i$, and whose unperturbed image is correctly classified. Note that since we are concerned with degree, the domain of the distribution function $\hat{\mathcal{B}}_i$ is a vector whose dimension is equal to the number of nodes in the induced graphs.  If for some input, the model outputs class $i$, we would like to know if the output was generated by a random variable with distribution $\mathcal{B}_i$ or with distribution $\mathcal{A}_i$ where the lack of a hat denotes the true distribution.As before, we first construct the graph $\mathcal{G}$ for the sample and compute a sample degree vector, which we denote by the random variable $\mathbf{Z}$.  For a yet to be defined subset of nodes $\mathcal{S}$, we define the following Wasserstein Sums Ratio (WSR) quantity:

\begin{equation}
\label{eq:test}
\text{WSR}(\mathcal{S}, \hat{\mathcal{A}}_i,\hat{\mathcal{B}}_i,\mathbf{Z},i)=  \frac{\sum_{j \in  \mathcal{S}}  \mathcal{W}(    \delta_{\mathbf{Z}_j},  \hat{\mathcal{B}}_i^j )  }{\sum_{j \in  \mathcal{S}}  \mathcal{W}(\delta_{\mathbf{Z}_j},   \hat{\mathcal{A}}_i^j  ) }
\end{equation}
where the $j$ in $\hat{\mathcal{A}}_i^j$ refers to the empirical distribution for node $j$, and similarly for $\hat{\mathcal{B}}_i^j$. Equation \eqref{eq:test} says that for each node that belongs to  $\mathcal{S}$, we compute Wasserstein-1 distances node-wise from the sample to the empirical distributions and we sum over the node indices and compute the ratio. If the ratio is less than some threshold, we classify the input as benign, otherwise as adversarial. It may occur that the denominator of \eqref{eq:test} is equal to 0, thus, in this case, a small term is added to the numerator and denominator. This can happen if the empirical distributions $ \{\hat{\mathcal{A}}_i^j\}_{j \in  \mathcal{S}} $  only have support at a point.  Lastly, we note that we could have also computed the Wasserstein distance in $\mathbb{R}^{N}$, where $N$ is the number of nodes in $\mathcal{G}$. However, that is a more involved procedure. Using \eqref{eq:wassestein:discrete}, we can write \eqref{eq:test}
as 

\begin{equation}
\label{eq:test2}
\text{WSR}(\mathcal{S}, \hat{\mathcal{A}},\hat{\mathcal{B}},\mathbf{Z},i)=  \frac{ \frac{1}{ N_{ \hat{\mathcal{B}}_i^j}} \sum_{j \in  \mathcal{S}}  \sum_{k=1}^{ N_{ \hat{\mathcal{B}}_i^j} } \|  \mathbf{Z}_j-  y_i^j(k) \|_1       }{ \frac{1}{ N_{ \hat{\mathcal{A}}_i^j}}   \sum_{j \in  \mathcal{S}}  \sum_{k=1}^{ N_{ \hat{\mathcal{A}}_i^j} } \|  \mathbf{Z}_j-  x_i^j(k) \|_1  }
\end{equation}
where  $y_i^j(k)$ is a sample from $\hat{\mathcal{B}}_i^j$ and $ x_i^j(k)$ is a sample from $\hat{\mathcal{A}}_i^j$, and  $N_{ \hat{\mathcal{B}}_i^j}$ is the number of samples in $\hat{\mathcal{B}}_i^j$, respectively for $\hat{\mathcal{A}}_i^j$. Lastly, we make the set $\mathcal{S}$ as follows: we calculate 
\begin{equation}
\label{eq:delta}
\Delta_i^j := \mathbb{E}  X_i^j  - \mathbb{E}  Y_i^j 
\end{equation}
where $X_i^j$ has distribution  $\hat{\mathcal{A}}_i^j$ and $Y_i^j$ has distribution $\hat{\mathcal{B}}_i^j$ and $ \mathbb{E}$ is expected value. We then create the set  
\begin{equation}
\mathcal{S} := \{ j: \Delta_i^j < 0  \ \text{for all} \ i \}.
\end{equation}
The  set $\mathcal{S}$ identifies nodes where the mean of the benign distribution is greater than the adversarial distribution for all classes. Should it happen that $\hat{\mathcal{A}}_i^j$ is empty for some $j$ (we have experienced this only for one combination of model and attack), one may create a placeholder version of it by setting each entry to a very large negative value (the large negative value has the effect of removing the index $j$ from consideration when making the set $\mathcal{S}$).
Algorithm~\ref{alg:orig} shows adversarial detection using WSR.

\begin{algorithm}[hbt!]
   \caption{Adversarial detection using WSR (variant 1)}
   \label{alg:orig}
    \begin{algorithmic}
       \State {\bfseries Input:} neural network $\mathcal{NN}$, $\text{image} \ I$; ${\tau}$, $\mathcal{S}$ , $\hat{\mathcal{A}}_i^j$; $\hat{\mathcal{B}}_i^j$ for all $i$ and $j$
       \State $i \gets \mathcal{NN}(I)$
       \State compute $\mathcal{G}$ from $I$ and $\mathcal{NN}$
       \State compute node degree $\mathbf{z}$ from $\mathcal{G}$
       \State $val \gets \text{WSR}(\mathcal{S}, \hat{\mathcal{A}},\hat{\mathcal{B}},\mathbf{z},i) $
       \State if $val$ \textless $\tau$ then classify $I$ as benign, otherwise classify $I$  as adversarial.
    \end{algorithmic}
\end{algorithm}

The way we construct $\mathcal{S}$ has the tendency to pick nodes that generalize well across all classes at the expense of nodes that specialize.  In an alternative algorithm, we propose to use the specialized nodes. For a given output that is classified as class $i$, we use $\mathcal{S}_i=\{j: \Delta_i^j < 0\}$. This can result in a more accurate test using our approach, but at the expense of a little longer computation since there are more nodes to use for computations. The algorithm is shown in Algorithm~\ref{alg:variant}.

\begin{algorithm}[tbh!]
   \caption{Adversarial detection using WSR (variant 2)}
   \label{alg:variant}
\begin{algorithmic}
   \State {\bfseries Input:} neural network $\mathcal{NN}$, $\text{image} \ I$; ${\tau_i}$, $\mathcal{S}_i$ , $\hat{\mathcal{A}}_i^j$; $\hat{\mathcal{B}}_i^j$ for all $i$ and $j$
   \State  $i \gets \mathcal{NN}(I)$
   \State  compute $\mathcal{G}$ from $I$ and $\mathcal{NN}$
   \State compute node degree $\mathbf{z}$ from $\mathcal{G}$
   \State $val \gets \text{WSR}(\mathcal{S}_i, \hat{\mathcal{A}},\hat{\mathcal{B}},\mathbf{z},i) $
   \State if $val$ \textless $\tau_i$ then classify $I$ as benign, otherwise classify $I$  as adversarial.
\end{algorithmic}
\end{algorithm}

\section{Consistency}
\label{sec:consistency}
We would like to analyze under what conditions \eqref{eq:test} is a faithful predictor.  We treat the case of a finite-width ANN with sufficiently many neurons. A finite-width ANN has the property that the degree distribution has compact support, which implies that the Wasserstein distance between an empirical degree distribution and true distribution is bounded, and the Wasserstein distance is continuous with respect to $\| \cdot\|_{\infty}$.   We begin our proof of consistency by showing that given a real-valued random variable $X$; an empirical distribution $\hat{F}_n$ of some other real-valued random variable with true distribution $F$; a function $G$ (whose arguments are a random variable and a distribution) that is uniformly continuous in the second argument with respect to  $\| \cdot\|_{\infty}$; and bounded, that
\begin{equation}
\label{eq:conv1}
   \mathbb{E}_X G(  X, \hat{F}_n) \xrightarrow{a.s.} \mathbb{E}_X G(   X,  F )
\end{equation}
as $n \to \infty$. To prove \eqref{eq:conv1}, it is sufficient to show that 
\begin{equation}
\label{eq:conv2}
 G(  X, \hat{F}_n) \xrightarrow{a.s.}  G(  X,  F) \ \forall{x}.
\end{equation}
Under identical and independently distributed (iid) assumptions, the Glivenko-Cantelli lemma states that $\| \hat{F}_n  - \hat{F} \|_{\infty} \xrightarrow{a.s.} 0$. This combined with the uniform continuity of $G$ in the second argument with respect to  $\| \cdot\|_{\infty}$ proves \eqref{eq:conv2}. To prove \eqref{eq:conv1}, we let $h_n(x)= G(  x, \hat{F}_n)$  and  $h(x)= G(  x, F)$. 
From \eqref{eq:conv2} we have $h_n(x) \xrightarrow{a.s.} h(x)$ for all $x$ as $n \to \infty$. We may combine this with the boundedness assumption to use the Lebesgue dominated convergence theorem, resulting in $\lim_{n \to \infty} \mathbb{E}_X h_n(X) =\mathbb{E}_X \lim_{n \to \infty}  h_n(X) =\mathbb{E}_X h(X)$ almost surely.

We now begin to analyze \eqref{eq:test}, and we start by supposing that our random variable $\mathbf{Z}$ corresponds to the benign case.  Let
\begin{align}
\begin{split}
\label{eq:x_b}
U_{j,i}^b &  = \mathcal{W}(    \delta_{\mathbf{Z}_j},  \hat{\mathcal{B}}_i^j ) \\
U_{j,i}^a &  = \mathcal{W}(    \delta_{\mathbf{Z}_j},  \hat{\mathcal{A}}_i^j ).
\end{split}
\end{align}
For additional simplicity, let us assume that quantities defined in \eqref{eq:x_b}  are iid over the index $j$. The iid assumption implies that 

\[  \mathbf{E}_{\mathbf{Z}_j} \mathcal{W}(    \delta_{\mathbf{Z}_j},  \hat{\mathcal{B}}_i^j )=: \mathbf{E} U_{i}^b\]
and
\[ \mathbf{E}_{\mathbf{Z}_j}  \mathcal{W}(    \delta_{\mathbf{Z}_j},  \hat{\mathcal{A}}_i^j ) =: \mathbf{E}  U_{i}^a \]
for all $i$. By equation~\eqref{eq:conv1}, the results we obtain going forward will hold for the population distribution in high probability assuming our empirical distributions have enough samples. By the weak law of large numbers,
\[ \left|\frac{\sum_{j=1}^{|\mathcal{S}|} \mathcal{W}(    \delta_{\mathbf{Z}_j},  \hat{\mathcal{B}}_i^j ) }{|\mathcal{S}|} - \mathbf{E} U_i^b   \right| < \epsilon_1 \ \text{as} \ |\mathcal{S}| \to \infty \]
Similarly,
\[  \left| \frac{ \sum_{j=1}^{|\mathcal{S}|} \mathcal{W}(    \delta_{\mathbf{Z}_j},  \hat{\mathcal{A}}_i^j )  }{ |\mathcal{S}|  } - \mathbf{E} U_i^a  \right| < \epsilon_2 \ \text{as} \ |\mathcal{S}| \to \infty.  \]  
Then \eqref{eq:test} is equal to

\begin{align}
\label{eq:large_num1}
\frac{\sum_{j=1}^{|\mathcal{S}|} U_{j,i}^b }{\sum_{j=1}^{|\mathcal{S}|} U_{j,i}^a } &=\frac{|\mathcal{S}| \mathbf{E} U_{i}^b+ |\mathcal{S}|\epsilon_1}{|\mathcal{S}| \mathbf{E} U_{i}^a+ |\mathcal{S}| \epsilon_2} \notag \\
&= \frac{ \mathbf{E} U_{i}^b+  \epsilon_1}{\mathbf{E} U_{i}^a+  \epsilon_2} \notag \\
& \to  \frac{ \mathbf{E} U_i^b}{\mathbf{E} U_i^a} \ \text{as } \ |\mathcal{S}| \to \infty
\end{align}
where $\epsilon_1$ and $\epsilon_2$ are $o(|\mathcal{S}|)$. If we consider the case when $\mathbf{Z}$ is adversarial, we get a similar limit as in \eqref{eq:large_num1}. Thus for consistency, we need the two limits to not be equal, thus we write

\begin{equation}
\label{eq:different-limits}
\frac{ \mathbf{E} U_i^b}{\mathbf{E} U_i^a} < \frac{ \mathbf{E} V_i^b}{\mathbf{E} V_i^a}
\end{equation}
where we use $V$ to denote adversarial quantities. This is equivalent to  $\mathbf{E} U_i^b \  \mathbf{E} V_i^a < \mathbf{E} U_i^a  \ \mathbf{E} V_i^b$. This is a realistic assumption for distributions with different means. A classification threshold, $\tau$, is then picked such that

\begin{equation}
\label{eq:different-limits-tau}
\frac{ \mathbf{E} U_i^b}{\mathbf{E} U_i^a} < \tau < \frac{ \mathbf{E} V_i^b}{\mathbf{E} V_i^a}.
\end{equation}
 An interesting example of  \eqref{eq:different-limits} is the case in which $\mathbf{E} U_i^b=  \mathbf{E} V_i^a$ and  $\mathbf{E} U_i^a=  \mathbf{E} V_i^b$ and where all terms do not equal 1. In this instance, \eqref{eq:test} in the benign case will be the inverse of that in the adversarial case. Furthermore, neither ratio will equal 1. This happens when adversarial distributions are simply shifts of benign distributions.

\section{Experimental details}

\subsection{Datasets}
We train our models on MNIST, CIFAR-10, and SVHN datasets. For each model we create adversarial examples using the Adversarial Robustness Toolbox \cite{art2018}.  For CIFAR-10 and SVHN, all images were enlarged to (224, 224, 3). Images are preprocessed using built-in Keras layers that handle input preprocessing.

\subsection{Architectures}
We experiment with five models, two of which are detailed in Tables~\ref{table:model1}-\ref{table:model2} while the other three are VGG-19, InceptionResnetV2 and MobileNet. The last layers of VGG-19, InceptionResnetV2, and MobileNet are preloaded from Keras, and their last layers are replaced with three custom, fully-connected layers, with output sizes 4096, 1000, and 10, respectively, and trained with ImageNet weights. With respect to these three models, we only compute graph-based quantities from these layers to keep the run-time reasonable. For models 1 and 2, we use all layers. 

\subsection{Hyperparameters}
The values of $\epsilon$ and $\alpha$ in our implementation of LRP-$\alpha \beta$ are 2 and $10^{-7}$, respectively. In our implementation of RSA we use $M=8, K=16$, and the layer used is the third from the output layer. For creating noisy samples in the algorithm in LID, we use Gaussian noise of zero mean and variance 0.05. Also in our implementation of LID, we only use the last 10 layers for computational ease.

\subsection{Attacks}
We consider the fast gradient sign method, \cite{goodfellow}, projected gradient descent \cite{madry2018}, untargeted Carlini-Wagner L2 \cite{carlini-wagner}, DeepFool \cite{moosavi}, Square \cite{maksym}, and Auto \cite{croce_auto} attacks. Fast gradient sign method attacks are clipped when perturbations are outside a ball of radius 10\% in the $\ell^{\infty}$ norm. Projected gradient descent attacks are crafted using the same norm but with up to a 5\% perturbation; the number of iterations was 40 except for InceptionResnetV2, MobileNet, and VGG19, in which 10 were used.  Square and Auto attacks have the same norm and perturbation as projected gradient descent attacks. Optimization is done using ADAM with learning rate 0.01. For each attack we generate 10,000 adversarially perturbed images from 10,000 original (test data) images. In creating training data for the detection methods we introduce, approximately 14,000 samples were used, and the methods are compared on approximately 6,000 samples. For RSA the numbers are approximately the same. For LID, we use approximately 6,000 training and test samples each, with the exception of models 1 and 2 in which we use approximately 7,000 training and 3,000 test samples as these models were simpler and thus enabled faster computations.

\begin{table}[t]
\caption{Architecture of Model 1} 
\label{table:model1}
\vskip 0.1in
\begin{center}
\begin{small}
\begin{sc}
\resizebox{\columnwidth}{!}{%
\begin{tabular}{@{}||c c c ||@{}} 
 \hline
 layer type  & output size & activation function  \\ [0.5ex] 
 \hline\hline
 fully connected & 300 & reLu  \\ 
 \hline
 fully connected & 200 & reLu\\
 \hline
fully connected & 150 & reLu \\
 \hline
 fully connected & 150 & reLu \\
 \hline
 fully connected & 100 & sigmoid  \\ 
  \hline
 fully connected & 10 & softmax \\ 
 \hline
\end{tabular}
}
\end{sc}
\end{small}
\end{center}
\vskip 0.1in
\end{table}

\begin{table}[t]
\caption{Architecture of Model 2} 
\label{table:model2}
\vskip 0.1in
\begin{center}
\begin{small}
\begin{sc}
\resizebox{\columnwidth}{!}{
\begin{tabular}{@{}||c c c ||@{}} 
 \hline
 layer type  & output size & activation function  \\ [0.5ex] 
 \hline\hline
 conv  & 3 filters, kernel size (4,4) & identity \\ 
 \hline
 maxpool & pool size=(2,2), strides=(2,2) & reLu\\
 \hline
 conv  & 3 filters, kernel size (4,4) & identity \\ 
 \hline
maxpool & pool size=(2,2), strides=(2,2) & reLu\\
 \hline
 fully connected & 100 & reLu  \\ 
  \hline
 fully connected & 10 & softmax \\ 
 \hline
\end{tabular}
}
\end{sc}
\end{small}
\end{center}
\vskip 0.1in
\end{table}

\section{Results and Discussion}

\subsection{Comparison of logistic regression approaches}
In Tables~\ref{table:benchmarking-logreg-cifar10}, \ref{table:benchmarking-logreg-svhn},and \ref{table:benchmarking-logreg-mnist} we report the  specificity (percentage benign samples that are correctly detected) and sensitivity (percentage adversarial samples that are correctly detected). One can see that the various graph statistics considered here can be strong, sensitive, and specific predictors of adversarial attacks in the case of using logistic regression. Among Mobilenet, InceptionResnetV2 and VGG19, the three graph-based statistics perform more or less the same in their classification of samples as benign or adversarial and they generally perform well. For instance, when MobileNet is given SVHN data that is attacked by Square, the specificity and sensitivity are 99.66\% and 99.04\%, respectively.  From the tables, we see that the worst performance tends to occur under Carlini-Wagner and Deepfool attacks. These two attacks are believed to be among the most difficult to detect, so our results are consistent with this assumption. In particular, when VGG19 is given SVHN data that is attacked by Carlini-Wagner, the results show that our logistic regression models are good at detecting benign samples, but mostly fail to catch the adversarial ones. To be more concrete, when considering degree, the specificity and sensitivity are 97.67\% and 29.68\%, respectively. For node importance, the values are 98.84\% and 9.68\%, respectively. When using outermost edge relevance, the corresponding values are 100\% and 0\%, meaning that our detector erroneously classifies all input as benign; apparently in this case, the differences between outermost edge relevances in the benign and adversarial cases are too similar.

Among Model 1 and 2, degree is a significantly better predictor than node importance and outermost edge relevance, while outermost edge relevance for Model 2 is a poor predictor across all attacks, being unable to detect adversarial images. This is because the outermost edge relevance for benign and adversarial samples are thresholded to 0. The largest relevances for Model 2 are found in layers closer to the input layer. 

In comparison to LID, our results are superior across almost all model/attack combinations. The tables show that for LID, the accuracy averaged between benign and adversarial detection rates is around 50\% which is much less than than what we achieve using our graph-based statistics. We note that \cite{drenkow} also reports similar numbers for implementation of LID, thereby concurring with our results. 

Lastly, in Figure~\ref{fig:intrinsic_graph_data}, we include sample data for the three graph statistics introduced in Section~\ref{sec:graph_generation}.  Note that because there are thousands of nodes, we only choose one node to show.

\begin{table}[!htb]
\centering
\caption{Comparison between logistic regression methods. First and second quantities in each entry are benign and adversarial detection rate, respectively. FGSM, PGD, CW2, DF, Sq, and AA represent fast gradient sign method, projected gradient descent, Carlini-Wagner L2, Square, and Auto attacks, respectively. Values are percentages. Outermost edge relevance for Model 2 is a poor predictor because the outermost edge relevance for benign and adversarial samples are identical (0). During the thresholding process, the relevance for the edges corresponding to the output layer are set to 0 because they are relatively small.} 
\vskip 0.1in
   \subfloat[CIFAR-10   \label{table:benchmarking-logreg-cifar10}]{
\resizebox{\columnwidth}{!}{%
\begin{tabular}{c|c|c|c|c|c|c|c|c|}
\cline{2-9}
\multirow{8}{*}[-2cm]{\rotatebox{90}{model}} & && \multicolumn{6}{c|}{\textbf{Attack}}\\ \cline{4-9}
 & &  & FGSM & PGD & CW2 & DF & Sq & AA\\
\cline{2-9}
 &\multirow{4}{2cm}{MobileNet}
 & degree  &  99.64/98.10  & 99.63/99.23 & 82.21/90.74 & 80.23/91.34 & 94.08/93.22 & 100/99.63 \\ \cline{3-9}
 && node importance   &  99.52/99.35  & 99.46/100 & 64.50/99.43 & 66.78/93.91 & 91.86/93.39 &  99.89/100 \\ \cline{3-9}
 &&  outermost edge relevance &  100/99.98  & 99.71/99.61 & 85.27/90.06 & 87.75/89.24 & 100/99.85 & 100/99.86 \\ \cline{3-9}
 &&   LID    & 23.49/85.93  & 12.01/97.90 & 22.30/79.84 & 35.59/68.10 & 24.47/98.60 & 7.02/87.35\\ \cline{2-9}
&\multirow{4}{4cm}{InceptionResNetV2}
 & degree      &  100/99.95  & 100/99.92 & 84.05/94.31 & 73.88/86.22 & 96.18/98.50& 99.66/100\\  \cline{3-9}
 && node importance   &  100/100  & 100/100 & 65.76/88.13 & 44.12/96.08 & 89.30/99.36 & 100/99.96 \\  \cline{3-9}
&& outermost edge relevance     &  99.96/99.70  & 100/99.70 & 80.36/78.76 & 80.96/74.43 & 99.21/96.59& 100/99.78 \\  \cline{3-9}
 &&   LID & 40.13/66.23   &  76.58/44.09  & 21.08/73.91 & 72.97/28.78 & 67.79/41.59 & 31.60/75.93\\ \cline{2-9}
&\multirow{4}{2cm}{VGG19}
 & degree     &  100/99.86  & 99.92/99.73 & 99.77/04.20 & 97.55/99.08 & 98.98/98.21& 99.16/100\\  \cline{3-9}
 && node importance       & 99.96/100   & 99.96/100 & 100/0.70  & 94.88/99.35 & 98.24/98.28 & 97.97/100\\  \cline{3-9}
&& outermost edge relevance      & 99.73/99.72 & 99.96/99.96 & 100/0  & 99.66/96.82 & 99.66/97.21 & 100/99.92 \\  \cline{3-9}
 &&   LID  & 79.66/6.57  & 99.39/42.71  & 41.77/57.55 & 99.73/0.88 & 98.78/0.11 & 22.14/76.08 \\ \cline{3-9}
\cline{2-9}
\end{tabular}
}
   }\\
   \vskip 0.1in
   \subfloat[SVHN \label{table:benchmarking-logreg-svhn}]{
\resizebox{\columnwidth}{!}{%
\begin{tabular}{c|c|c|c|c|c|c|c|c|}
\cline{2-9}
\multirow{8}{*}[-2cm]{\rotatebox{90}{model}} & && \multicolumn{6}{c|}{\textbf{Attack}}\\ \cline{4-9}
 & &  & FGSM & PGD & CW2 & DF & Sq & AA\\
\cline{2-9}
&\multirow{4}{2cm}{MobileNet}
 & degree     &  100/100  & 100/99.77 & 77.25/91.87 & 55.44/92.96 & 99.66/99.04 & 100/100 \\ \cline{3-9}
 && node importance   &  100/100  & 99.66/100 & 55.61/90.53 & 54.39/88.69 & 89.30/99.36 & 100/99.96  \\ \cline{3-9} 
 && outermost edge relevance &  100/99.79  & 100/99.85 & 99.59/76.24 & 79.85/74.90 & 99.21/96.96 & 100/99.81 \\ \cline{3-9}
  &&   LID    &  57.12/40.29  & 69.78/37.14 & 77.47/23.03 & 92.97/7.03 & 88.45/7.71& 85.23/37.52\\ \cline{2-9}
&\multirow{4}{4cm}{InceptionResNetV2}
 & degree      &  99.88/99.84  & 98.63/100 & 76.00/91.52 & 75.29/92.63 & 87.35/97.11 & 99.41/100\\  \cline{3-9}
 && node importance   &  100/100  & 100/99.96 & 45.68/93.81 & 89.52/94.36 & 89.52/94.36 & 100/100\\  \cline{3-9}
&& outermost edge relevance     &  99.96/99.68  & 100/99.60 & 74.06/76.25 & 92.95/90.58 & 92.95/90.58 & 100/99.92 \\  \cline{3-9}
 &&   LID   & 92.14/4.10   & 69.78/37.14  & 77.17/23.02& 53.32/48.55 & 29.87/59.27 & 10.75/61.80 \\ \cline{2-9}
&\multirow{4}{2cm}{VGG19}
 & degree     &  100/99.93  & 100/100 & 97.67/29.68 & 98.80/99.18 & 96.97/99.57 & 99.96/100\\  \cline{3-9}
 && node importance  & 100/99.94   & 100/99.96 & 98.84/9.68  & 99.34/99.39 & 98.64/99.73 & 100/100\\  \cline{3-9}
&& outermost edge relevance      & 100/99.88 & 100/99.81 & 100/0  & 99.96/98.37 & 100/97.83 & 100/99.92 \\  \cline{3-9}
 &&   LID  & 67.79/40.38  & 2.26/85.83  & 42.24/47.81 & 4.14/92.37 & 7.01/98.72 & 6.47/29.00 \\ \cline{3-9}
\cline{2-9}
\end{tabular}
}
   }\\
   \vskip 0.1in
   \subfloat[MNIST \label{table:benchmarking-logreg-mnist}]{
\resizebox{\columnwidth}{!}{%
\begin{tabular}{c|c|c|c|c|c|c|c|c|}
\cline{2-9}
\multirow{8}{*}[-1cm]{\rotatebox{90}{model}} & && \multicolumn{6}{c|}{\textbf{Attack}}\\ \cline{4-9}
 & &  & FGSM & PGD & CW2 & DF & Sq & AA\\
\cline{2-9}
&\multirow{4}{2cm}{Model 2}
 & degree     &  99.00/95.45  & 99.90/95.45& 98.05/99.19 & 97.26/99.20 & 99.86/65.44 & 99.97/95.34 \\ \cline{3-9}
 && node importance   &  84.80/24.54 &  92.79/12.46 & 52.75/62.37 & 34.22/73.53 & 97.71/5.53 & 94.50/8.04  \\ \cline{3-9}
 && outermost edge relevance &  100/0  & 100/0 & 100/0 & 100/0 & 100/0 & 100/0 \\ \cline{3-9}
&&   LID    & 81.94/13.43  & 88.19/10.41 & 74.90/21.72 & 71.79/22.27 & 52.77/50.96 & 16.10/87.53 \\\cline{2-9}
&\multirow{4}{2cm}{Model 1}
 & degree      &  95.96/98.91  & 94.76/84.19& 6.58/88.76 & 95.51/96.71 & 98.63/71.97 & 96.02/84.48\\  \cline{3-9}
 && node importance   &  81.67/96.19  & 96.09/54.83 & 0.14/99.70 & 88.14/96.57 & 100/2.09 & 93.87/60.39 \\  \cline{3-9}
&& outermost edge relevance     &  90.34/86.50  & 89.31/85.25 & 49.55/42.19 & 95.68/93.58 & 100/0 & 89.47/88.57  \\  \cline{3-9}
 &&   LID    & 50.60/49.75 &   72.07/25.38 & 47.20/55.53 & 71.79/22.27 & 52.77/50.96 & 26.71/72.12\\ \cline{3-9}
\cline{2-9}
\end{tabular}
}
   }
\vskip 0.1in
\end{table}

\clearpage

\begin{center}
\captionsetup{type=figure}
\vskip 0.2in
\begin{subfigure}[b]{.32\columnwidth}
    \centering
    \includegraphics[width=1\columnwidth]{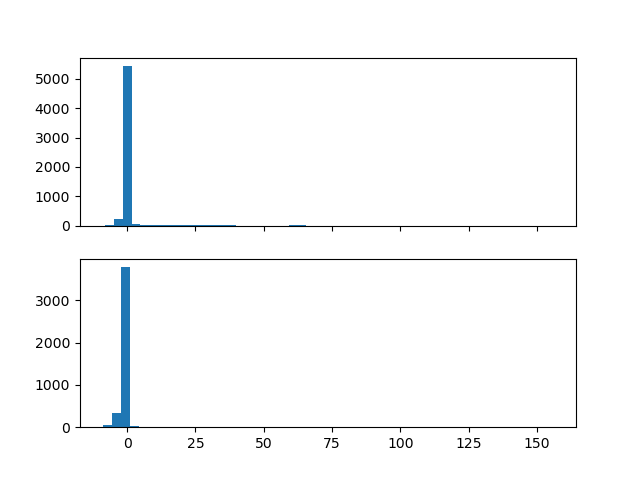}
    \caption{Top subfigure is a histogram of degree for benign images, computed for node number 426. Bottom subfigure shows the histogram computed for adversarial images.}
\end{subfigure}
\hfill
\begin{subfigure}[b]{.32\columnwidth}
    \includegraphics[width=1\columnwidth]{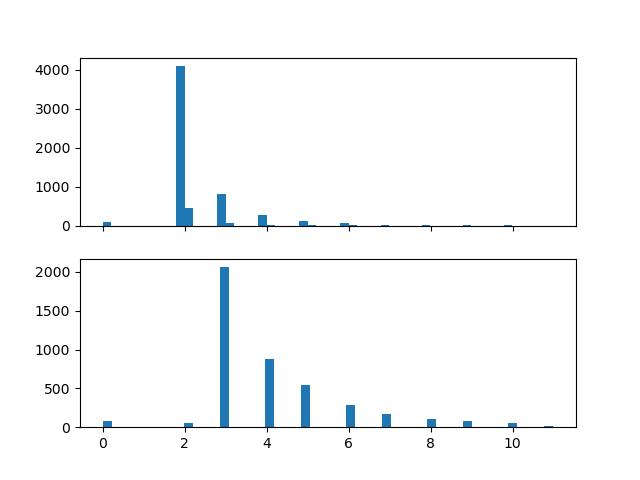}
    \caption{Top subfigure is a histogram of node importance for benign images, computed for node number 426. Bottom subfigure shows the histogram computed for adversarial images.}
\end{subfigure}
\hfill
\begin{subfigure}[b]{.32\columnwidth}
    \includegraphics[width=1\columnwidth]{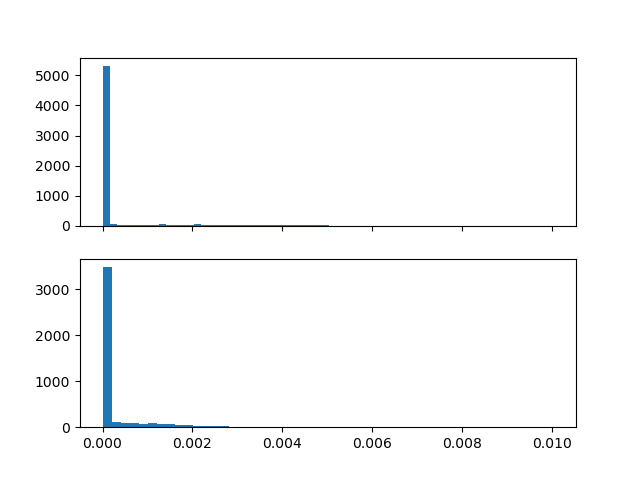}
    \caption{Top subfigure is a histogram of outermost edge relevance for benign images, computed for a certain edge. Bottom subfigure shows the histogram computed for adversarial images.}
\end{subfigure}
\captionof{figure}{Data shown is for InceptionResNetV2, Square attack.}
\label{fig:intrinsic_graph_data}
\end{center}

\subsection{ Comparison of statistical approaches}

Tables~\ref{table:benchmarking-wsr-cifar10}, \ref{table:benchmarking-wsr-svhn}, and \ref{table:benchmarking-wsr-mnist} show results in terms of AUROC (area under receiver operating characteristic curve) for various detection methods. In almost all cases, WSR2 provides more accurate predictions than WSR1 as predicted in Section~\ref{sec:consistency}. Further, both WSR variations outperform RSA. Model 1, in comparison to the other models, performs somewhat poorly under WSR1. This seems to be due to Model 1 having the least number of neurons, making the corresponding $|\mathcal{S}|$ relatively small which implies slower convergence of the statistical test. More generally, we can also see from the tables that model/attack pairs with small $|\mathcal{S}|$ tend to have worse results under WSR. This is particularly noticeable in the case of Carlini-Wagner and Deepfool attacks under WSR1; this lower performance was also noted in our results using logistic regression, and provides evidence in support of the belief that Carlini-Wagner and Deepfool attacks are among the strongest. In Figure~\ref{fig:empirical_B_A }, one can see histograms of the distributions $\hat{\mathcal{B}}_i^j$ and $\hat{\mathcal{A}}_i^j$ for $j=426$, $i=3$. Because there are thousands of nodes, we only choose one value of $j$ to show.

\begin{table}[H]
\centering
\caption{Comparison of AUROC for statistical detection methods. FGSM, PGD, CW2, DF, Sq, and AA represent fast gradient sign method, projected gradient descent, Carlini-Wagner L2, Deepfool, Square, and Auto attacks, respectively. WSR1 and WSR2 are WSR variants 1 and 2 respectively.  Values are percentages.} 
\vskip 0.1in
   \subfloat[CIFAR-10   \label{table:benchmarking-wsr-cifar10}]{
\resizebox{\columnwidth}{!}{%
\begin{tabular}{c|c|c|c|c|c|c|c|c|}
\cline{2-9}
\multirow{8}{*}[-1.5cm]{\rotatebox{90}{model}} & && \multicolumn{6}{c|}{\textbf{Attack}}\\ \cline{4-9}
 & &  & FGSM & PGD & CW2 & DF & Sq & AA\\
\cline{2-9}
&\multirow{3}{2cm}{MobileNet}
 & WSR1     &  99.98  & 100 & 79.78 & 73.56 & 92.98 & 100\\  \cline{3-9}
 && WSR2     &  100  & 94.47 & 98.00 & 92.11 & 99.25 & 100\\  \cline{3-9}
 && RSA       & 51.31  & 68.56 & 84.94  & 48.79 & 97.44 & 85.74 \\  \cline{2-9}
&\multirow{3}{4cm}{InceptionResNetV2}
 & WSR1   & 99.64 & 99.00 & 81.46 & 64.23 & 84.88 & 100 \\  \cline{3-9}
 && WSR2   & 97.26 & 99.05 & 95.69 & 88.54 & 89.93 & 99.98 \\  \cline{3-9}
 && RSA   & 94.67  & 98.89 & 99.06 & 98.59  & 98.72 & 95.07 \\  \cline{2-9}
 &\multirow{3}{2cm}{VGG19}
 & WSR1   & 100  & 99.98 & 74.83 & 98.78 & 97.08 & 99.98\\  \cline{3-9}
  && WSR2   & 99.87  & 99.97 & 97.36 & 99.32 & 99.70 & 100\\  \cline{3-9}
 && RSA      & 69.08  & 54.28 & 71.70 & 73.77 & 76.58 & 63.78\\  \cline{3-9}
\cline{2-9}
\end{tabular}
}
   }\\
   \vskip 0.1in
   \subfloat[SVHN \label{table:benchmarking-wsr-svhn}]{
\resizebox{\columnwidth}{!}{%
\begin{tabular}{c|c|c|c|c|c|c|c|c|}
\cline{2-9}
\multirow{8}{*}[-1.5cm]{\rotatebox{90}{model}} & && \multicolumn{6}{c|}{\textbf{Attack}}\\ \cline{4-9}
 & &  & FGSM & PGD & CW2 & DF & Sq & AA\\
\cline{2-9}
&\multirow{3}{2cm}{MobileNet}
 & WSR1     & 100  & 100 & 81.62 & 78.95 & 99.29 & 100\\  \cline{3-9}
  && WSR2     & 100  & 100 & 95.20 & 90.26 & 99.80 & 100\\  \cline{3-9}
 && RSA     & 80.12  & 63.12 & 87.69  & 82.83 & 76.48 & 82.27 \\  \cline{2-9}
&\multirow{3}{4cm}{InceptionResNetV2}
 & WSR1   & 100 & 100 & 78.35 & 80.59 &92.12 & 100 \\  \cline{3-9}
&& WSR2   & 99.86 & 99.91 & 93.84 & 94.29 & 96.81 & 100 \\  \cline{3-9}
 && RSA   & 51.77  & 61.81 & 56.35 & 60.25 & 58.44 & 58.19 \\  \cline{2-9}
&\multirow{3}{2cm}{VGG19}
 & WSR1   & 100 & 100 & 76.33 & 99.65 & 99.54 & 100\\  \cline{3-9}
  && WSR2   & 100 & 100 & 94.57 & 99.68 & 99.87 & 100\\  \cline{3-9}
 && RSA   & 80.69  & 57.27 & 76.79 & 79.63 & 81.77 & 58.23\\  \cline{2-9}
\cline{2-9}
\cline{2-9}
\end{tabular}
}
   }\\
   \vskip 0.1in
   \subfloat[MNIST \label{table:benchmarking-wsr-mnist}]{
\resizebox{\columnwidth}{!}{%
\begin{tabular}{c|c|c|c|c|c|c|c|c|}
\cline{2-9}
\multirow{8}{*}[-.5cm]{\rotatebox{90}{model}} & && \multicolumn{6}{c|}{\textbf{Attack}}\\ \cline{4-9}
 & &  & FGSM & PGD & CW2 & DF & Sq & AA\\
\cline{2-9}
&\multirow{3}{2cm}{Model 2}
 & WSR1   & 95.16  & 95.37 & 96.48  & 95.44 & 91.77 & 95.54 \\  \cline{3-9}
  && WSR2   & 96.75  & 96.23 & 96.73  & 96.08 & 91.74 & 96.68 \\  \cline{3-9}
 && RSA  & 66.81  & 62.81 & 58.54 & 55.95 & 68.33 & 63.12\\  \cline{2-9}
&\multirow{3}{2cm}{Model 1}
 & WSR1  & 95.53  & 81.06 & 41.25 & 82.48 & 89.30 & 83.60 \\  \cline{3-9}
  && WSR2  & 96.36  & 94.40 & 39.65 & 97.81 & 99.48 & 94.36 \\  \cline{3-9}
 && RSA  & 71.80  & 72.10 & 51.64 & 89.60 & 96.34 & 71.85\\  \cline{2-9}
\cline{2-9}
\end{tabular}
}
   }
\vskip 0.1in
\end{table}

\begin{center}
\captionsetup{type=figure}
\vskip 0.2in
    \includegraphics[width=0.6\columnwidth]{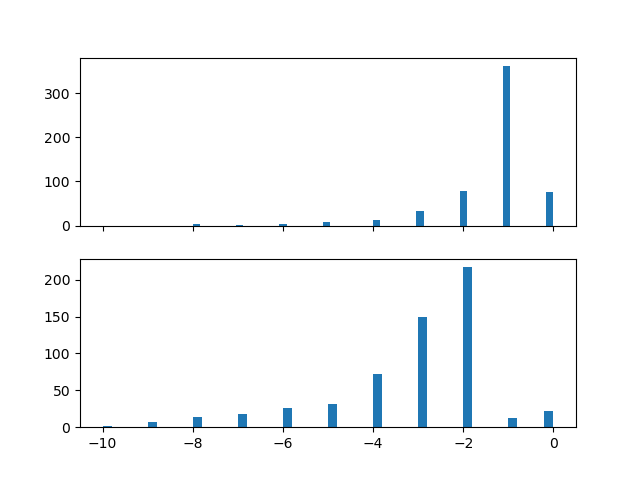}
\caption{Data shown is for Inceptionresnetv2, Square attack. Top subfigure shows histogram of $\hat{\mathcal{B}}_i^j$ for $j=426$, $i=3$. Bottom subfigure shows histogram of $\hat{\mathcal{A}}_i^j$ for $j=426$, $i=3$. }
\label{fig:empirical_B_A }
\end{center}

\begin{center}
\vskip 0.2in
\captionsetup{type=figure}
\begin{subfigure}[b]{\columnwidth}
    \centering
    \includegraphics[width=0.15\columnwidth]{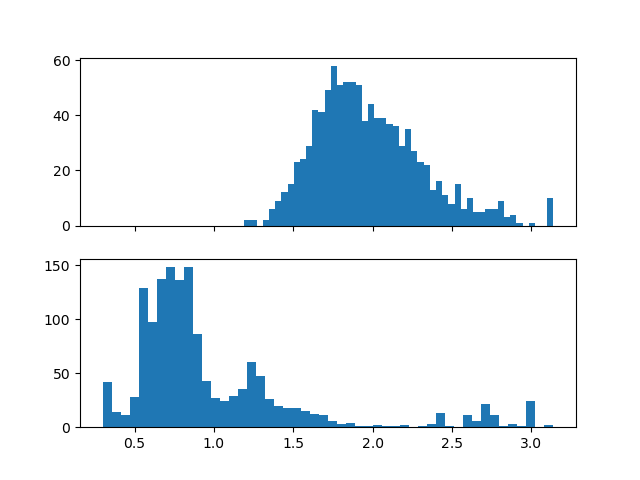}
    \centering
    \includegraphics[width=0.15\columnwidth]{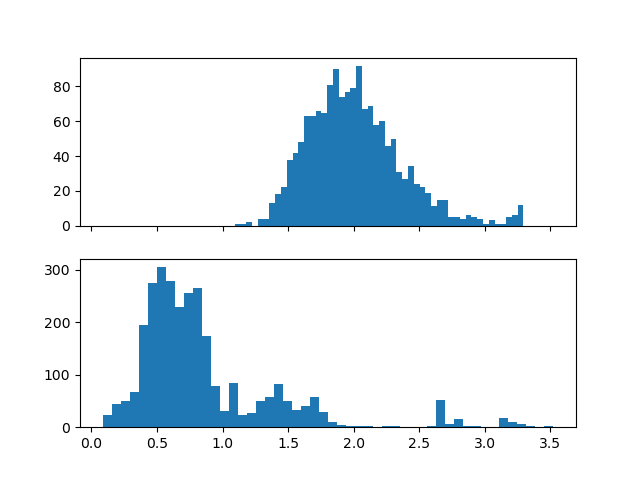}
    \centering
    \includegraphics[width=0.15\columnwidth]{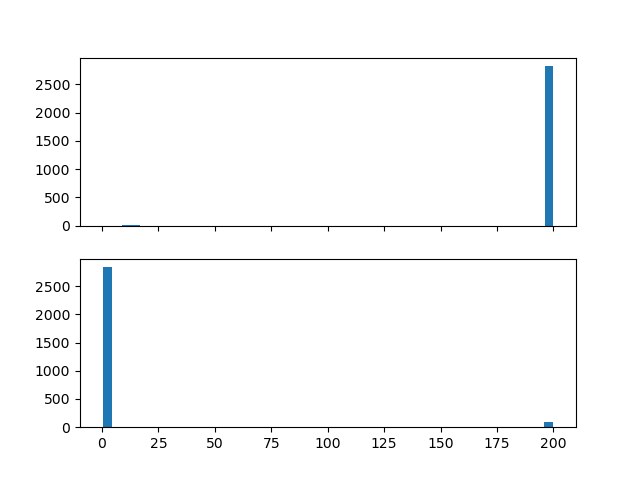}
    \centering
    \includegraphics[width=0.15\columnwidth]{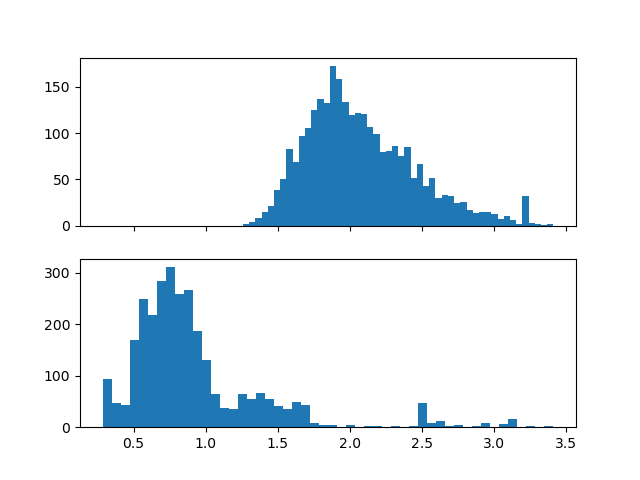}
    \centering
    \includegraphics[width=0.15\columnwidth]{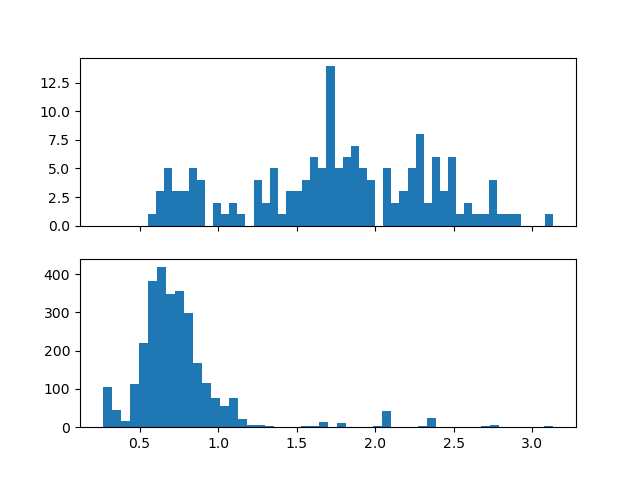}
    \centering
    \includegraphics[width=0.15\columnwidth]{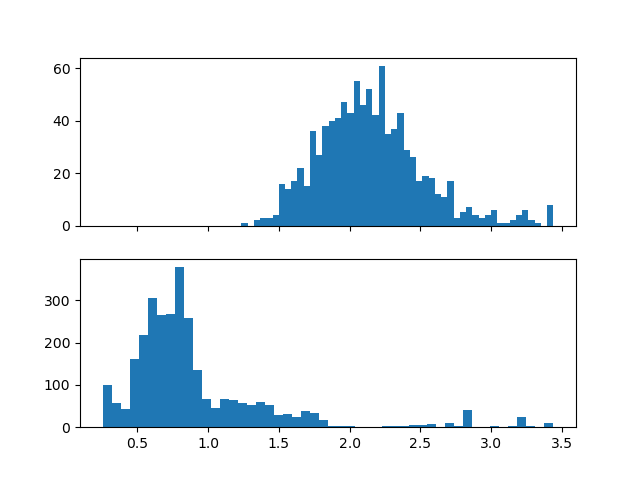}
    \caption{WSR1 for Model 2 }
\end{subfigure}
\begin{subfigure}[b]{\columnwidth}
    \centering
    \includegraphics[width=0.15\columnwidth]{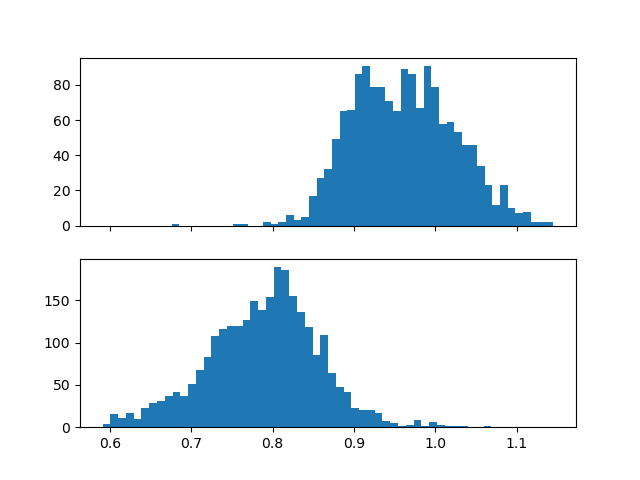}
    \centering
    \includegraphics[width=0.15\columnwidth]{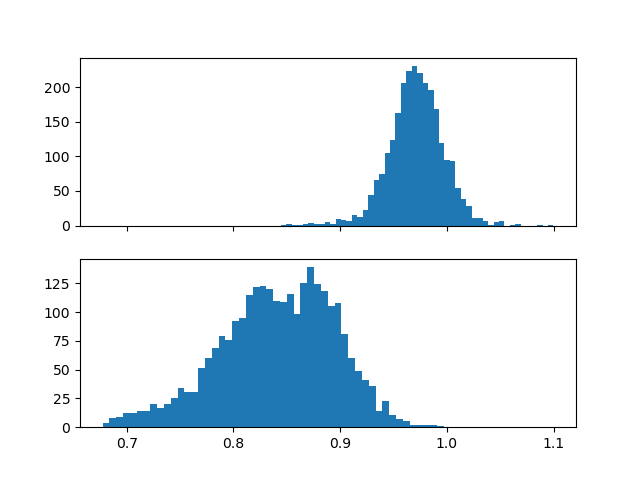}
    \centering
    \includegraphics[width=0.15\columnwidth]{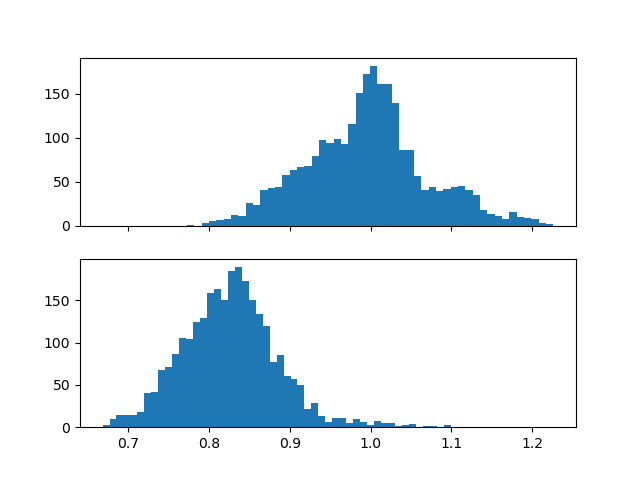}
    \centering
    \includegraphics[width=0.15\columnwidth]{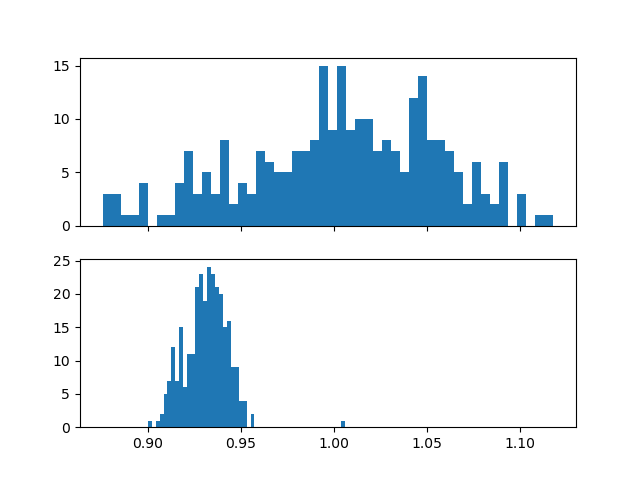}
    \centering
    \includegraphics[width=0.15\columnwidth]{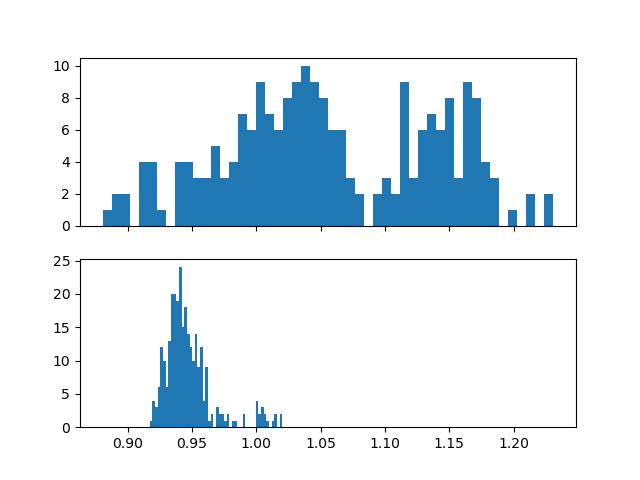}
    \centering
    \includegraphics[width=0.15\columnwidth]{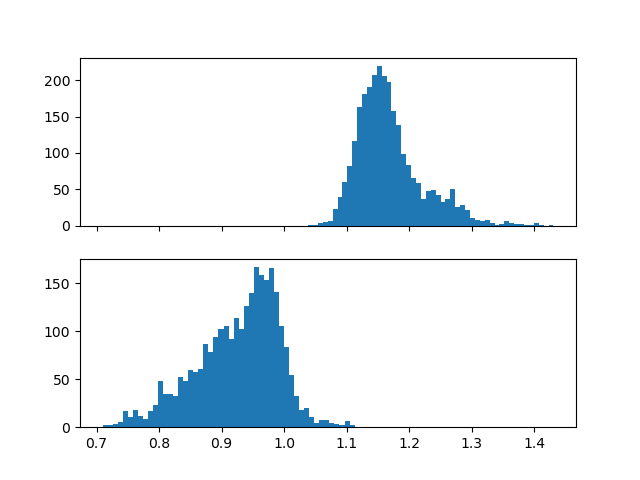}
    \caption{WSR2 for InceptionResnetV2. }
\end{subfigure}
\caption{Empirical distributions of WSR1 for Model 2 and WSR2 for InceptionResnetV2. The top subplot of a figure shows WSR computed for adversarial examples, and the bottom subplot shows the computation for benign examples. FGSM, PGD, CW2, DF, Sq, and AA represent fast gradient sign method, projected gradient descent, Carlini-Wagner L2, Deepfool, Square, and Auto attacks, respectively. For Model 2 and CW2, values above 200 are set to 200 for ease of display. Note that the benign and adversarial plots for Model 2 tend to agree with the remark made in Section~\ref{introduction} about inverses. }
\label{fig:dist_WR}
\vskip 0.2in
\end{center}

\clearpage

We can use WSR and logistic regression in a complementary way. For instance, graph-based quantities generated from VGG19 and Carlini-Wagner attacks tend to be poorly classified with logistic regression. In contrast, WSR2 performs well in this case, and it can be used in place of logistic regression.

As an aside, in \cite{roth19}, the authors claim that adding noise to images affects the logits in such a way that adversarial inputs can be detected. We considered using equation 5 from \cite{roth19} as a baseline in place of RSA, but chose not to because of the extremely large time needed for the source code to run and secondly, our initial results suggested that this method gives poor accuracy, near 50\%, which is much lower than the performance reports in the literature. In our effort to increase accuracy we experimented with different hyperparameters, including noise, but to no avail. This calls into question the usefulness and robustness of using equation 5 in \cite{roth19}.

\subsection{Nodal analysis}
The distribution of node quantities is highly dependent on the model and attack. It can be seen (Table 5) that AUROC for $\text{WSR}$ decreases as the strength of the attack increases (we consider a partial order of in increasing attack strength to be: Fast Gradient Sign Method, Projected Gradient Descent and Carlini-Wagner L2). We can relate this observation to how the cardinality of $|\mathcal{S}|$ varies with model/attack. The cardinality of $|\mathcal{S}|$ can be seen in Tables~\ref{table:S-cifar10}, \ref{table:S-svhn}, and \ref{table:S-mnist}. For CIFAR-10 and SVHN datasets, we observe that the cardinality tends to be a lot smaller for Carlini-Wagner L2 and DeepFool attacks, and it seems to explain the lower accuracy achieved by WSR on these attacks. We recall that from Section~\ref{sec:consistency}, the accuracy of WSR increases with $|\mathcal{S}|$.

We also note that in some cases the benign distribution of $\text{WSR}$ and the adversarial distribution of $\text{WSR}$ are centered at points which are close to inverses. This seems to be the case for Model 2, as shown in Figure~\ref{fig:dist_WR}. This is in agreement with an earlier remark in Section~\ref{sec:consistency} about equation~\eqref{eq:test} having inverse values under benign and adversarial examples.

\begin{table}[!htb]
\centering
\caption{Cardinality of  $\mathcal{S}$  by model and attack. FFGSM, PGD, CW2, DF, Sq, and AA represent fast gradient sign method, projected gradient descent, Carlini-Wagner L2, Deepfool, Square, and Auto attacks, respectively.}
   \subfloat[CIFAR-10   \label{table:S-cifar10}]{
\resizebox{\columnwidth}{!}{%
\begin{tabular}{c|c|c|c|c|c|c|c|c|}
\cline{2-9}
\multirow{8}{*}[-.0cm]{\rotatebox{90}{model}} & && \multicolumn{6}{c|}{\textbf{Attack}}\\ \cline{4-9}
 & &  & FGSM & PGD & CW2 & DF & Sq & AA\\
\cline{2-9}
&\multirow{1}{4cm}{MobileNet}
 & WSR1   & 877  & 734 & 51 & 68 & 405 & 79 \\  \cline{2-9}
&\multirow{1}{4cm}{InceptionResnetV2}
 & WSR1   &378  & 484 & 109 & 180 & 240 & 120 \\ \cline{2-9}
&\multirow{1}{4cm}{VGG19}
 & WSR1   & 180  & 776 & 34 & 925 & 692 & 578  \\ \cline{2-9}
\end{tabular}
}

   }\\
   \vskip 0.1in
   \subfloat[SVHN \label{table:S-svhn}]{
\resizebox{\columnwidth}{!}{%
\begin{tabular}{c|c|c|c|c|c|c|c|c|}
\cline{2-9}
\multirow{8}{*}[-.0cm]{\rotatebox{90}{model}} & && \multicolumn{6}{c|}{\textbf{Attack}}\\ \cline{4-9}
 & &  & FGSM & PGD & CW2 & DF & Sq & AA\\
\cline{2-9}
&\multirow{1}{4cm}{MobileNet}
 & WSR1   &  422  & 636 & 78 & 97 & 551 & 287 \\  \cline{2-9}
&\multirow{1}{4cm}{InceptionResnetV2}
 & WSR1    &  754  & 1269 & 109 & 160 & 922 & 496\\ \cline{2-9}
&\multirow{1}{4cm}{VGG19}
 & WSR1  & 945 & 1253 & 63 & 1379 & 929 & 843  \\ \cline{2-9}
\end{tabular}
}
   }\\
   \vskip 0.1in
   \subfloat[MNIST \label{table:S-mnist}]{
\resizebox{\columnwidth}{!}{%
\begin{tabular}{c|c|c|c|c|c|c|c|c|}
\cline{2-9}
\multirow{8}{*}[.5cm]{\rotatebox{90}{model}} & && \multicolumn{6}{c|}{\textbf{Attack}}\\ \cline{4-9}
 & &  & FGSM & PGD & CW2 & DF & Sq & AA\\
\cline{2-9}
&\multirow{1}{4cm}{Model 2}
 & WSR1   & 129  & 105 & 50 & 34 & 52 & 115 \\  \cline{2-9}
&\multirow{1}{4cm}{Model 1}
 & WSR1    & 33  & 10 & 7  & 5 & 4 & 10\\ \cline{2-9}
\end{tabular}
}
   }
\vskip 0.1in
\end{table}

\section{Conclusion}

We demonstrate that neural network architectures can be interpreted in a graph context from which we can use the statistics of graph-based quantities to detect adversarial attacks. We introduce three statistics that we applied to our graphs and used them as predictors of adversarial attack. We show that this approach can produce high detection performances with logistic regression. We also study the distributions of node degree using a statistical test based on Wasserstein distances. One limitation of our method is that layer-wise relevance propagation is designed for ANNs with ReLU activations. Thus, the use of layer-wise relevance propagation may produce less explainability for neural networks with non-ReLu activations. However, ReLU is arguably the most popular type. It would be interesting to expand the class of neural networks in this study to include those with more non-ReLu activations, despite the possible gap in explainability. We note that Model 1 includes one non-ReLu activation (sigmoid) and it can be considered a small step in this direction.  Another direction of our work may include the use of other saliency map methods such as \cite{tobias} or \cite{simonyan} instead of layer-wise relevance propagation. Further, an ablation study can be done by removing specific and sensitive nodes from a neural network. In our context, these nodes can be from the set $\mathcal{S}$ that is used in our statistical tests. It would be interesting to see the relationship between the neural network's classification accuracy and the effectiveness of the statistical tests. We find it intriguing that a sparse graph encodes sufficient information about inputs to a neural network.  We hope that the perspective introduced here will provide a different perspective of understanding adversarial attacks.

\section{Acknowledgments}
L. Carboni and D. Nwaigwe are the recipients of a grant from MIAI@Grenoble Alpes (ANR 19-P3IA-003).

\clearpage 

\bibliographystyle{apalike}
\bibliography{bibliography.bib}

\end{document}